\documentclass{article}

\usepackage{microtype}
\usepackage{graphicx}
\usepackage{subfigure}
\usepackage{booktabs} 

\usepackage{hyperref}

\usepackage[accepted]{icml2020}

\icmltitlerunning{Towards User Friendly Medication Mapping Using Entity-Boosted Two-Tower Neural Network}

\begin{document}

\twocolumn[
\icmltitle{Towards User Friendly Medication Mapping Using Entity-Boosted Two-Tower Neural Network}



\icmlsetsymbol{equal}{*}

\begin{icmlauthorlist}
\icmlauthor{Shaoqing Yuan}{amz}
\icmlauthor{Parminder Bhatia}{aws}
\icmlauthor{Busra Celikkaya}{aws}
\icmlauthor{Haiyang Liu}{amz}
\icmlauthor{Kyunghwan Choi}{amz}
\end{icmlauthorlist}

\icmlaffiliation{aws}{Amazon AWS, Seattle, Washington, USA}
\icmlaffiliation{amz}{Amazon Alexa, Seattle, Washington, USA}

\icmlcorrespondingauthor{Shaoqing Yuan}{shaoqiny@amazon.com}

\icmlkeywords{Machine Learning, ICML}

\vskip 0.3in
]



\printAffiliationsAndNotice{}  
\begin{abstract}
Recent advancements in medical entity linking have been applied in the area of scientific literature and social media data. However, with the adoption of telemedicine and conversational agents such as Alexa in healthcare settings, medical name inference has become an important task.  Medication name inference is the task of mapping user friendly medication names from a free-form text to a concept in a normalized medication list. This is challenging due to the differences in the use of medical terminology from health care professionals and user conversations coming from the lay public.  We begin with mapping descriptive medication phrases (DMP) to standard medication names (SMN). Given the prescriptions of each patient, we want to provide them with the flexibility of referring to the medication in their preferred ways. We approach this as a ranking problem which maps SMN to DMP by ordering the list of medications in the patient’s prescription list obtained from pharmacies. Furthermore, we leveraged the output of intermediate layers and performed medication clustering. We present the Medication Inference Model (MIM) achieving state-of-the-art results. By incorporating medical entities based attention, we have obtained further improvement for ranking models. 
\end{abstract}

%

\section{Introduction}

\begin{figure}[htp]
    \centering
    \includegraphics[width=8cm]{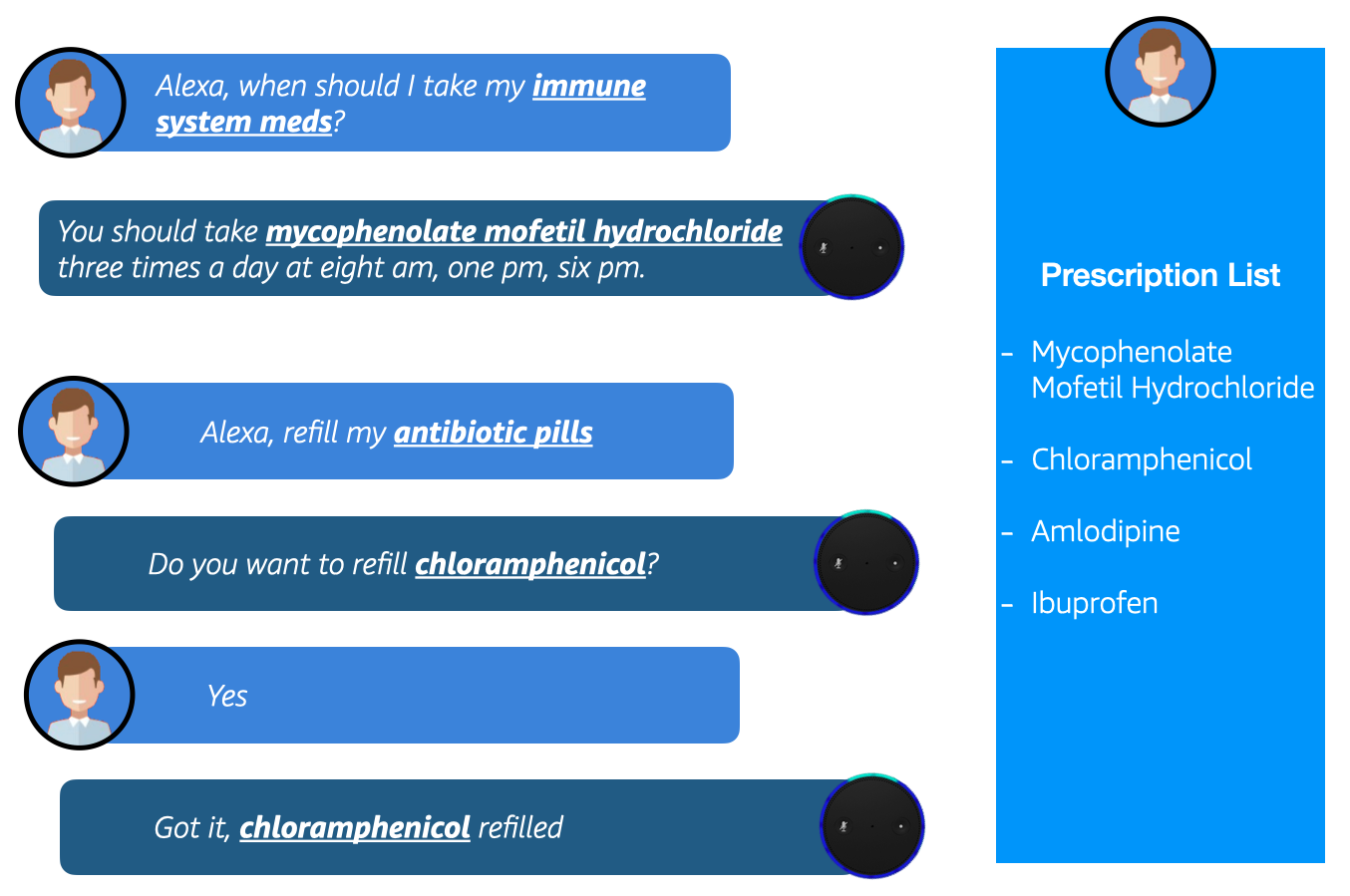}
    \caption{Different ways of how users interact with conversational agents for medical queries}
    \label{fig:BertPerformance}
\end{figure}

Medication names are extremely hard to pronounce for patients without a proper medical background. Thus, when interacting with Alexa on medication names, patients without this background may have many different ways to refer to a medication (e,g. Bumetanide can be referred to as Bumetanide (generic name), Bumex (brand name), high blood pressure pill (disease name)). On the other hand, patients with medical knowledge may use abbreviations or specialized ways to refer to medication names. For example, patients may use “immune meds” to refer to “mycophenolate mofetil hydrochloride” in their prescription list.

In this paper, we describe a new problem about finding the generic medication name (SMN: standardized medication name) based on a patient's description (DMP: descriptive medication phrase) from a list of medications the patient is consuming. According to our internal user research, in the United States, patients with chronic diseases usually take around four to five medications daily. This problem is different from medical concept normalization \citep{zhu2019latte} which tries to map a health-related entity mention in a free-form text to a concept in a controlled vocabulary \citep{miftahutdinov2019deep} which is a generic concept list rather than a patient specific prescription list and is generally much longer. 

We structure this  as a ranking problem. Here we rank all medications a patient is consuming based on the relationship with the patient's description and the one ranked highest will be the inference result.  We present a hard attention based entity boosted CNN architecture achieving 4\% over earlier ranking methods. 

Furthermore, the mapping between SMN and DMP contains the patient's understanding of the medications, especially from the usage perspective of the medications. Using latent output from our model, we build a medication clustering system which groups together medications with similar effects and disease treatments. The output is designed to aid physicians to consider other medications as a substitution for decreasing cost as well as helping patients distinguish medications that are similar in their impression but should, in reality, be used in different conditions. Moreover, with clustering patients will have an intuitive understanding of the relationship of the medications they are consuming. 
Our contributions are as follows:

\begin{itemize}
    \item We present a medical entity boosted architecture, Medication Inference Model (MIM) achieving a $7\% - 9\%$ improvement over strong BERT baselines.
    
    \item We benchmark against state-of-the-art ranking architectures,  demonstrating robustness of our work.
    
    \item We present medication clustering results which group together medications with similar effects and treat the similar diseases.
\end{itemize}

\section{Task Definition}
Each example is represented as a tuple $(Q, P_1, P_2, ..., P_n, Y)$, where $Q = (q_1, q_2, ..., q_{l_q})$ is a DMP, with a length $l_q$, $P_i = (p_1,  p_2, ...p_{l_{p_i}})$ is a SMN, with a length $l_{p_i}$, and $Y = (y_1, y_2, ..., y_n)$ is the label representing the relationship between $Q$ and $P_1, P_2, ..., P_n$. $Y$ and $P$ have the same length. $y_i = 1$ if $P_i$ is the generic medication name that $Q$ is referring to, $0$ otherwise. 

\begin{table}[ht]
    \centering
    \begin{small}
    \begin{tabular}{ll}
    \toprule
    
    $Y$	& $Q$ = 	\{$q_1$(high), $q_2$(blood), 	$q_3$(pressure)\} \\
    $y_1$ = $0$	& $P_1$ = 	\{$p_1$(morphine),  $p_2$(suppository)\} \\
    $y_2$ = $1$	& $P_2$ = 	\{$p_1$(hydrochlorothiazide)\}		\\
    $y_3$ = $0$	& $P_3$ = 	\{…	\}		\\
    …		& …			\\
    
    \bottomrule
    \end{tabular}
    \label{table:data example}
    \end{small}
    \vspace*{-5mm}
\end{table}

It is possible that among $P_1, P_2, ..., P_n$, more than one medications may be referred to by $Q$.  Thus, $\sum_{i=0}^{n}y_i = m$ where $m$ is the number of medications in $P_1, P_2, ..., P_n$ that could be referred by $Q$. Ideally, we should make it possible that for the estimated $\hat{Y}$, $\sum_{i=0}^{n}\hat{y_i} > 1$. In this paper, however, we assume $\sum_{i=0}^{n}y_i = 1$.

The clustering task is defined as grouping medications across the prescriptions of different patients. i.e., we assign each medication in $(P_1, P_2, ..., P_N)$ to a group according to the DMP $Q$ associated with them. 

\begin{figure*}[htp]
    \centering
    \includegraphics[width=12cm]{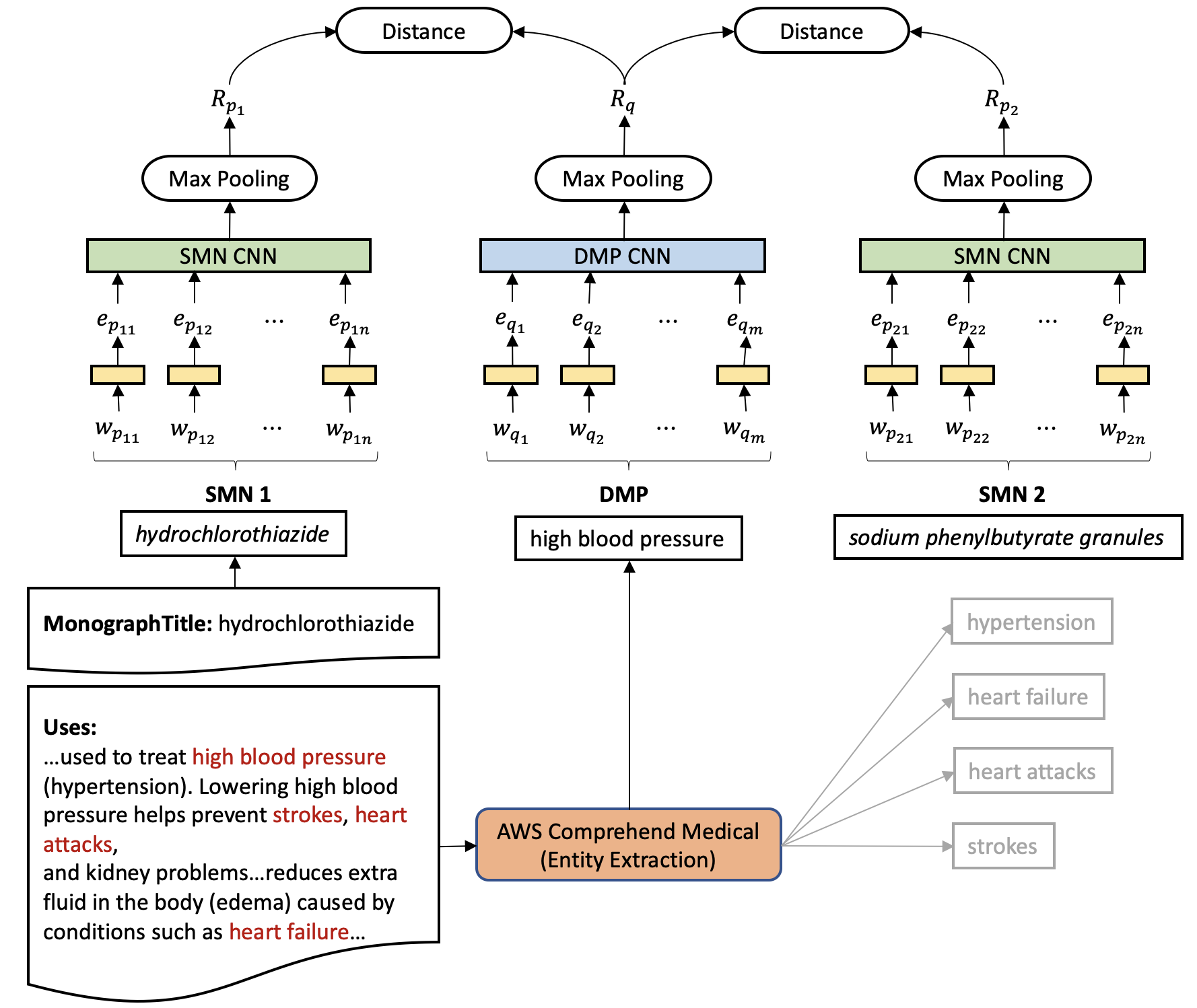}
    \caption{Medication Inference Model structure}
    \label{fig:SiameseStructure}
\end{figure*}

\section{Method}
Instead of comparing $n$ medications in each sample, we begin with two. Each sample consists of $(Q, P_1, P_2, Y)$ and the model must distinguish which SMN in ${ P_1, P_2}$ the patient is referring to by $Q$. To simplify, we assume only one of $( P_1, P_2)$ could be referred by $Q$. In practice, when there are $n$ (where $n > 2$) medications in a patient's prescription we run the model on all the combinations of the medications and rank them accordingly. 

\subsection{Entity Boosted Two-Tower Neural Network}
Motivated by the facial recognition problem, where the models evaluate the similarity between images of faces \citep{chopra2005learning}, we apply a Two-tower neural network to our problem. We regard the DMP as the query of each medicine and the SMNs as the medication candidates. The purpose is to match the correct SMNs to the DMP provided. 

Descriptions can often be verbose and can contain a large amount of noise. To improve the robustness and reduce noise, we have incorporated medical entity based hard attention \citep{luong2015effective} using Amazon Web Services Comprehend Medical (CM) \citep{8999113} which is a natural language processing service to perform entity and relation extraction.  

For each instance of data, we use the generic name as the SMN in our model, and generate DMPs from free-form text data that describe the usage of each medication in patient friendly terms. To reduce the noise,  we feed the description to CM to extract entities. CM is able to extract relevant medical information from unstructured text and classify extracted entities into five categories and 28 types. In this work we use the entities marked with types ``dx name'' (diagnostic indicator), ``treatment name'', ``system organ site'', ``swap'', ``generic name'', ``procedure name'', ``brand name'', ``test name'' as DMPs.

\subsubsection{Medication Inference Model} \label{Medication Inference Model}
Figure \ref{fig:SiameseStructure} outlines the Medication Inference Model (MIM) network.
In this model, there are two different sets of unshared embedding weights, where one is used to embed SMNs and the other DMP. 
We use convolutional (CNN) layers \citep{kim-2014-convolutional} followed by pooling on top of the embedding layer to get a vector representation of SMNs and DNP separately. 
We use cosine distance to measure the separation between the two SMNs and DMP. 

\subsubsection{BERT-based Model}
It is natural to leverage models which prove to be successful in solving question-answering problems to process our task. Here, DMP are regarded as the patient's query, with the $n$ SMNs as the answer candidates. 

We concatenate the DMP output with different SMNs separately and combine them into a BERT \citep{devlin2018bert} based multi-choice model \citep{Guo:2019:MLP:3331184.3331403}. The vector representations of the \textsf{\scriptsize [CLS]} tokens are used to represent the combinations of DMP with each SMN, and are fed into a fully connected layer. 
Finally we use hinge loss for ranking to compare the scale values with ground truth. 

\section{Experiments}

\begin{table*}[ht]
    \centering
    \begin{small}
    \begin{tabular}{lrrrr|rrrr}
    \toprule
      \multicolumn{9}{c}{\textit{Number of candidates}}  \\
      \multicolumn{5}{c|}{Baseline} &
      \multicolumn{4}{c}{Entity-based Attention} \\
    Model & 2 & 3  & 4 & 5 & 2 & 3  & 4 & 5   \\
    \midrule
    NormalBert & 69.7	& 54.7	& 46.6	& 42.7	& 79.7	& 72.7	& 62.4	& 62.5 \\
    BioBert & 73.4	& \textbf{59.8}	& \textbf{51.9}	& 46.4	& 81.5	& 74.8	& 66.2	& 66.4 \\
    ClinicalBert & 73.5 &	57.8 &	\textbf{51.9} &	47.4 & 79.5	& 74.4	& 64.8	& 62.7 \\
    ARC-1 & 65.2	& 46.2	& 37.5	& 33.4	& 76.8	& 66.8	& 58.6	& 59.4 \\
    ARC-2 & 64.0 &	48.1 &	39.1 &	36.5 & 75.7	& 65.7	& 62.1	& 57.0 \\
    ConvKNRM & 65.4 &	52.0 &	42.6 &	38.8 & 76.7	& 66.6	& 60.7	& 57.3 \\
    MatchLSTM & 64.9 &	50.8 &	40.9 &	35.4 & 82.4	& 70.9	& 62.6	& 58.3 \\
    MatchPyramid & 59.5 &	45.0 &	36.5 &	33.5 & 74.0	& 58.0	& 58.0	& 54.1 \\
    MIM & \textbf{73.9}	& 57.9	& 48.9	& \textbf{49.7}	& \textbf{87.7}	& \textbf{80.8}	& \textbf{78.9}	& \textbf{76.9} \\
    
    \bottomrule
    \end{tabular}
    \caption{Synthetic test set: For each model we report top1 accuracy including and excluding entity-based attention.}
    \label{table:results_synthetic}
    \end{small}
    \vspace*{-5mm}
\end{table*}

\begin{table}[ht]
    \centering
    \begin{small}
    \begin{tabular}{lrrrrr}
    \toprule
      \multicolumn{6}{c}{\textit{Upper limit for number of candidates}}  \\
    Model & 2 & 3  & 4 & 5 & 10  \\
    \midrule
    BioBert  & 83.10	& 73.50	& 67.00	& 61.8 &  52.70 \\
    MIM  & \textbf{83.50}	& \textbf{74.50}	& \textbf{69.50}	& \textbf{63.60} & \textbf{53.80} \\

    \bottomrule
    \end{tabular}
    \caption{Real test set: For each model we report top1 accuracy.}
    \label{table:results_real}
    \end{small}
    \vspace*{-5mm}
\end{table}

\subsection{Dataset and Evaluation Metrics} \label{Dataset and Evaluation Metrics}

\textbf{Synthetic Data Set} The training and test dataset are generated from 2,683 medication descriptions from the FDB\footnote{https://www.fdbhealth.com/} PEM (patient education module) dataset. FDB stands for First DataBank which is a known drug database and medical device database provider. Each PEM file contains a patient facing medication description including medication generic name, uses, warnings side effects etc.\footnote{CommonNames, Warning, Uses, HowToUse, SideEffects, Precautions, DrugInteractions, Overdose, Notes, MissedDose, Storage, MedicAlert}. 

The SMNs are collected from the generic name section and DMPs are generated from the "USES" section of the PEM files using CM as described in Section \ref{Medication Inference Model}. To evaluate the effect of the entity extraction component, we generate another DMP set by randomly drawing n-grams (where $n = [1,2,3,4,5]$) from the "USES" section of the PEM files as a replacement for CM. 

Next, we use the SMNs and DMPs collected to generate our training and test sets. we generate each instance starting with a DMP according to following steps. 

\begin{enumerate}
  \item For each DMP, we generate a positive SMNs set which consist of SMNs extracted from the same PEM file where the DMP is extracted from. It is possible that one DMP may have multiple positive SMNs if the DMP is a very general phrase. For example, the DMP ``high blood pressure'' may have multiple SMNs since many medications can be used to treat hypertension.

  \item For each DMP, we also generate a negative SMN set. The negative SMNs are all the medications covered by the  PEM files excluding the positive SMNs identified above and should follow the constraint that the entities extracted from the "USES" section of the negative SMNs' PEM files should have no overlap with that of the DMPs.
  
  \item Each instance in the training and validation data set consists of $1$ positive SMN and $n-1$ negative SMNs randomly selected from the SMN sets described above. The label of each instance is indicating which SMN is positive. 
\end{enumerate}

For the training and validation splits, $n$ is set to 2 in step 3 above, which means there are two SMNs in each instance. The  training and validation data set contains 680K instances and $70\%$ of them are used for training and $30\%$ for validation and testing.

For testing purposes, we generated four synthetic test sets with $n$ in step 3 set to $2, 3, 4, 5$ separately to simulate the real situations where patients with chronic disease in the U.S. usually have four to five medications in their prescription list at a time.

\textbf{Real Data Set} The real data set is generated based on 251 prescriptions collected from the i2b2 data set \footnote{https://portal.dbmi.hms.harvard.edu/projects/n2c2-nlp/} which contains the de-identified patient discharge summaries. Internal human annotators generate DMPs for each medication in the prescriptions. It is observed that in a real prescription, multiple medications may serve the same purpose and a general DMP could be used to refer to multiple medications in a prescription. In our current experiment, we assume the ground truth of each DMP is only the medication used to generate the DMP in a prescription. In this way, we will get the lower bound of the performance of the models. For testing purpose, we limited the number of medications in each prescription to be 10, 5, 4, 3, 2 respectively. For the test set with 10 as max number of medications, we go through all the 251 prescriptions and only select the prescriptions that has less than 10 medications into our test set. We randomly truncate the prescriptions in the 10 medication test set to 5, 4, 3, 2 medications as the other test sets. Further more, in order to evaluate the situations where one DMP may refer to multiple SMNs in a prescription, the annotators are currently working on labeling all the SMNs that a DMP could refer to in a prescription and if the model outputs one of the medications in the ground truth SMNs, the test sample will be marked as success in future experiments.

We report accuracy as the main evaluation metric, i.e., the correctness of selecting the positive SMN from $n$ SMNs. When evaluating on the test data, the model goes through all pairwise combinations of the SMNs and ranks all the SMNs accordingly.

\subsection{Experimental Details}
For the CNN-based model, we test multiple word embedding models including 200-dimensional BioWordVec \citep{zhang2019biowordvec, DBLP:journals/corr/abs-1810-09302} and 300-dimensional FastText word embeddings \citep{bojanowski2017enriching} trained with 3,466 articles from the Mayo Clinic. The two dimensional CNN layer consisted of 200 filters with window size 2, strip as 1 and no
regularization. Batch size is set to 150 and we observed model convergence after six epochs.  
For the pre-trained language model, we leverage Clinical BERT \citep{alsentzer-etal-2019-publicly}, BioBERT \citep{10.1093/bioinformatics/btz682}, and original BERT
models \citep{devlin2018bert}. We used the default settings for all BERT models as provided by \citet{devlin2018bert}. Batch size is set to 32, learning rate is set to $5 \times 10^{-5}$ and dropout rate is set to $0.2$. We observed the model converged after 10 epochs.  
We trained and evaluated all the models using a Tesla V100 GPU. 

\subsubsection{Baselines}
When evaluating the performance of our model, we compare the medication name inference performance with baseline models listed below. 

\begin{itemize}
    \item \textbf{ARC-I \citep{hu2014convolutional}}: ARC-I finds
the representation of each sentence with CNN layers, and then compares the representation for the two sentences with a multi-layer perceptron (MLP). 
    \item \textbf{ARC-II \citep{hu2014convolutional}}: ARC-II improves based on ARC-I by calculating the interaction features between sentences with CNN. 
    \item \textbf{ConvKNRM \citep{dai2018convolutional}}: Conv-KNRM uses CNN to represent n-grams of various lengths and soft matches them in a unified embedding space. The n-gram soft matches are then utilized by the kernel pooling and a fully connected layer to generate the final ranking score.
    \item \textbf{MatchLSTM \citep{wang2016machine}}: The matchLSTM sequentially aggregates the matching of the attention-weighted question to each token of the answer and uses the aggregated matching result to make a final prediction.
    \item \textbf{MatchPyramid \citep{pang2016text}}: MatchPyramid generates a matching matrix which represents the similarity between mention and candidate and then apply CNN layers on top of the matrix followed by a MLP layer to calculate the similarity score.
\end{itemize}

\begin{table*}[htp]
\begin{tabular}{|l|l|l|l|}
\hline
                              & \multicolumn{1}{c|}{\textbf{(a) Diagnose}}                                                                                                                               & \multicolumn{1}{c|}{\textbf{(b) Symptom}}                                                                                            & \multicolumn{1}{c|}{\textbf{(c) Drug type}}                                                                                                                  \\ \hline
\textit{\textbf{DMP}}         & \begin{tabular}[c]{@{}l@{}}high blood pressure, \\ strokes, heart attacks\end{tabular}                                                                                   & cough, coughing                                                                                                                      & antibiotic                                                                                                                                                   \\ \hline
\textit{\textbf{example SMN}} & amlodipine                                                                                                                                                               & promethazine                                                                                                                         & chloramphenicol                                                                                                                                              \\ \hline
\textit{\textbf{nearby SMNs}}  & \begin{tabular}[c]{@{}l@{}}perindopril, ramipril, \\ trandolapril, quinapril, \\ enalapril, isradipine, \\ lisinopril, sacubitril, \\ aliskiren, eplerenone\end{tabular} & \begin{tabular}[c]{@{}l@{}}dextromethorphan,\\ guaifenesin, expectorant,\\ antihistamine, \\ acetaminophen, hydrocodone, \\ zanamivir\end{tabular} & \begin{tabular}[c]{@{}l@{}}polymyxin b, gentamicin,\\ cefotetan, spiramycin,\\ gatifloxacin, piperacillin,\\ cephalexin, cefoxitin,\\ ofloxacin\end{tabular} \\ \hline
\end{tabular}
\caption{Examples of the DMP/SMN match and clustering results}
\label{table:clusterexample}
\end{table*}

\section{Results and Discussion}
Table \ref{table:results_synthetic} provides the accuracy results for each model we experimented with on the synthetic test data set. Number of candidates represent test data sets with $2,3,4,5$ medications in each test instance, as described in Section \ref{Dataset and Evaluation Metrics}. We report test results for each model with and without AWS Comprehend Medical as ``Entity-based Attention'' and ``Baseline'' columns.

Table \ref{table:results_synthetic} demonstrates the robustness of MIM. We observe that MIM and BERT based models outperform current state-of-the-art models such as ARC-I and MatchPyramid across a different number of candidates. Table \ref{table:results_real} further compares the performance of two best performing models on real test set. 

MIM outperforms BERT based models with a $7$-$9\%$ improvement in accuracy. 
We believe the major reason for this is that MIM, by encoding SMN and DMP separately, is able to encode the representation in a more robust way in comparison to BERT based models which concatenate the representations together using a special separator token.
We also observe performance variation of BERT models based on their pre-training. We found that domain specific pre-training helps, giving $2$-$3\%$ improvement when compared to the baseline BERT.

We observe the entity-boosted description gives robust results across all the model settings achieving significant improvement in accuracy over non-entity based models.
This alleviates the problem of noise in the lengthy descriptions.

Furthermore we see that our MIM model, with a relatively simpler CNN encoder as well as separate encoders for SMN and DMP, has the distinct advantage of generating inference results with low latency.  
This is ideal for real-time industrial settings. According to our experiments, the average latency for the MIM model for five medications is $10$ms, while, compared against BERT at $89$ms.

\subsection{Medication Clustering Result}
\begin{figure}[htp]
     \centering
     \includegraphics[width=7cm]{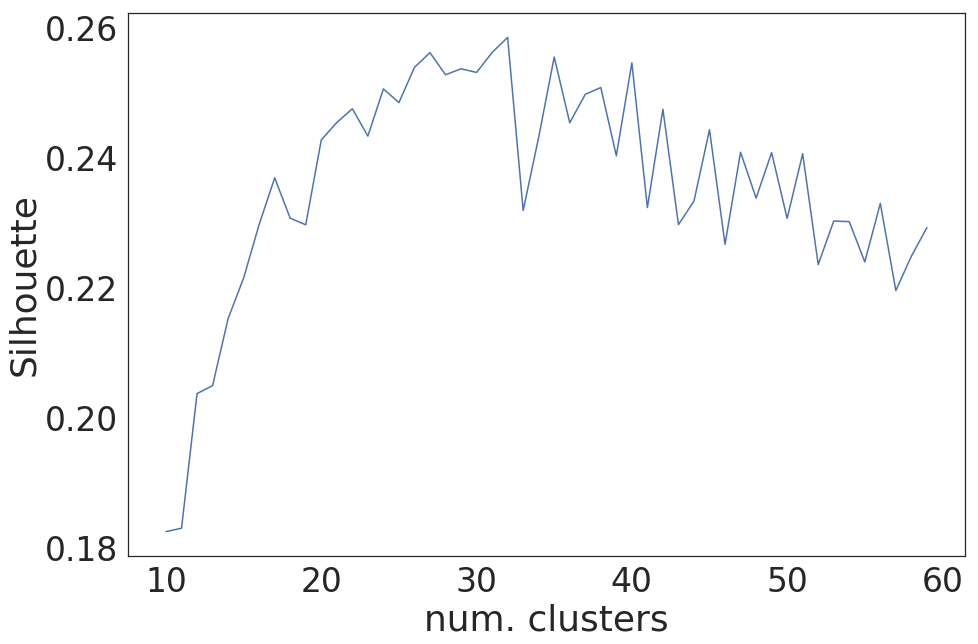}
     \caption{Average Silhouette value for different number of clusters}
     \label{fig:silhouette}
 \end{figure}

We apply \textit{k}-nearest neighbor (KNN) clustering based on the CNN max pooling output from the Two-tower neural network. 
The generic names of 2,683 medications are represented by vectors of 200 dimension. The number of clusters of KNN is determined by a Silhouette analysis\citep{rousseeuw1987silhouettes} with result given in Figure \ref{fig:silhouette}. The Silhouette analysis shows that clustering performs better when number of classes is 31.

\begin{figure*}[htp]
    \centering
    \includegraphics[width=16cm]{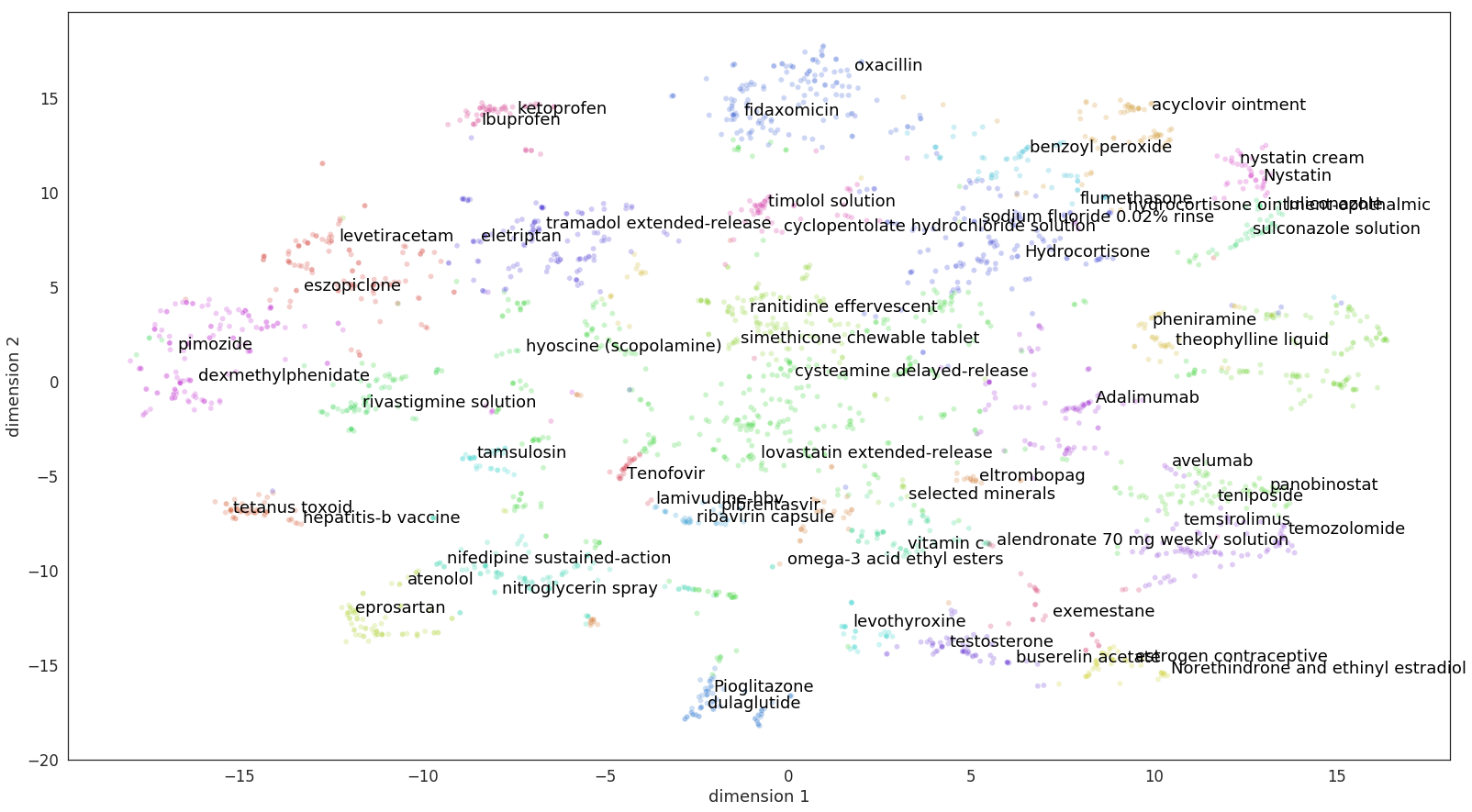}
    \caption{t-SNE visualization of the clusters}
    \label{fig:clustering}
\end{figure*}

Figure \ref{fig:clustering} shows the t-SNE visualization of the result where 2,683 medications are grouped into 31 clusters. The figure illustrates that medications with same effects, treating same disease or having similar drug types are mapped close to each other.

Table \ref{table:clusterexample} shows three examples of the clustering result. The nearby SMNs are sampled from the same cluster which example SMNs belong to and ranked according to their distance to the given SMN. The nature of the problem enables and requires the inference model to group SMNs based on multiple dimensions. We list examples from three dimensions and it is very natural for users to refer to their medications by diagnosis, disease symptoms and drug type. For example, coughing is a common symptom for multiple diseases including the common cold, pulmonary diseases such as pneumonia, and even from seasonal allergies. In column (b), the model is able to cluster medications that could relieve cough symptom of different underlying causes, for example, promethazine and antihistamine are used to treat allergies whereas zanamivir is used to treat and prevent flu.

\section{Related Work}
Earlier work on  medical concept normalization \citep{zhu2019latte} relied on lexicon based string matching and dictionary lookup to map limited number of variations of text to a pre-defined medical vocabulary \citep{aronson2001effective, brennan2003towards}. \citep{leaman2013dnorm} introduced DNrom as the first pairwise learning ranking model that compares associations between mentions and entities of various disease. \citep{limsopatham2016normalising, lee2017medical} then further leveraged deep learning models, convolutional neural network \citep{limsopatham2016normalising} and recurrent neural network models \citep{belousov2017using} trained on large corpus of medical articles etc.. Currently, researchers enhanced the deep learning based model with different model structure to incorporate context information, better process out of vocabulary (OOV) words and take advantages of interaction features from different semantic levels \citep{luo2018multi, miftahutdinov2019deep, niu2019multi}. 

With the success of deep learning, many neural network based models have been proposed for semantic matching, and document ranking. Models such as ARC-I \cite{hu2014convolutional} first compute the representation of the two sentences, and then compute their relevance. Semantic/text matching techniques fits well to solve the medical concept normalization problems when the number of candidates is limited. As listed in \cite{Guo:2019:MLP:3331184.3331403}, recently researchers have focused on developing deep learning models to solve document retrieval, question answering, conversational response ranking, and paraphrase identification \cite{Guo:2019:MLP:3331184.3331403} problems and introduced state-of-the-art models such as ARC-I \citep{hu2014convolutional}, ARC-II \citep{hu2014convolutional}, ConvKNRM \citep{dai2018convolutional}, MatchLSTM \citep{wang2016machine}, MatchPyramid \citep{pang2016text}, Bert\citep{devlin2018bert}. 

In recent years, natural language processing (NLP) techniques have demonstrated increasing effectiveness in clinical text mining  \citep{8999113} \citep{bhatia2019dynamic} . Electronic health record (EHR) narratives, e.g., discharge summaries and progress notes contain a wealth of medically relevant information such as diagnosis information and adverse drug events. Automatic extraction of such information and representation of clinical knowledge in standardized formats \cite{singh2019relation} could be employed for a variety of purposes such as clinical event surveillance, decision support \cite{jin2018improving}, pharmacovigilance, and drug efficacy studies.

This paper describes a problem that is a combination of the medical concept normalization and semantic matching problem using medical entity based hard attention. The nature of the problem presented in this paper requires the solution be able to extract informations from short phrases with limited context information.

\section{Conclusion and Future Work}
In this paper, we introduce a new problem common in the development of medication voice interaction products.
We evaluate the accuracy of different solutions and show that our entity boosted MIM outperform baseline models.  The specialty of this problem is that the context information is very limited when compared against other NLP tasks and the short length of the phrases prevent us from leveraging other advanced techniques that rely on words relationship in a phrase. The evaluation result also show that the problem prefers simple model structure. 
Since the phrases structure is very simple, the quality of word embeddings is more important in this problem and keeping the embedding weight unchanged is important when the training data is not sufficient enough to enhance the relationship between words either due to the nature of the data or small sample sizes. 

We also observe the discrepancy between synthetic collected datasets from real patients. For example, the combinations of the medicine on synthetic prescriptions may not be valid from a practitioner's or patient's perspective. We plan to further validate our model on real patient data to increase practicality. Finally on top of comparing and evaluating on two medications samples, we plan to experiment with more medications in each sample in training to closer mimic real world scenarios. 

\nocite{langley00}

\bibliography{example_paper}
\bibliographystyle{icml2020}


\end{document}